\documentclass[10pt,a4paper,twocolumn]{article}
\usepackage[T1]{fontenc}
\usepackage[utf8]{inputenc}
\usepackage{amsmath}
\usepackage{palatino}
\usepackage[charter]{newtxmath}
\usepackage[left=18mm,right=18mm,top=20mm,bottom=21mm]{geometry}
\usepackage{microtype,graphicx,booktabs,tabularx,array,mathtools}
\usepackage[authoryear,round]{natbib}
\usepackage{caption,enumitem,xcolor,titlesec}
\usepackage{placeins}
\usepackage{stfloats}
\usepackage{balance}
\usepackage{cuted}
\usepackage[colorlinks=true,linkcolor=blue,citecolor=blue,urlcolor=blue,breaklinks]{hyperref}
\usepackage{xurl}
\hypersetup{pdftitle={Adaptive DHiSparse: Speculative Decoding over a Load-Adaptive KV Hierarchy},pdfauthor={Naji Alazhar and Wassel Alazhar},pdfsubject={Long-context serving with adaptive residency, DSpark, and DFlash2}}
\DeclareCaptionLabelSeparator{bar}{\space|\space}
\setlist[itemize]{topsep=4pt,itemsep=3pt,parsep=0pt,leftmargin=1.6em}
\titleformat{\section}{\fontsize{11.5}{14}\selectfont\raggedright\hyphenpenalty=10000\bfseries}{\thesection.}{0.55em}{}
\titleformat{\subsection}{\normalsize\bfseries}{\thesubsection.}{0.55em}{}
\titleformat{\subsubsection}{\normalsize\bfseries}{\thesubsubsection.}{0.55em}{}
\titleformat{\paragraph}[runin]{\normalsize\bfseries}{}{0pt}{}[.]
\titlespacing*{\section}{0pt}{12pt plus 2pt minus 2pt}{5pt}
\titlespacing*{\subsection}{0pt}{10pt plus 2pt minus 1pt}{4pt}
\titlespacing*{\subsubsection}{0pt}{9pt plus 2pt minus 1pt}{4pt}
\titlespacing*{\paragraph}{0pt}{7pt plus 2pt minus 1pt}{0.7em}
\newcommand{\method}{Adaptive DHiSparse}
\newcommand{\Topk}{\operatorname{Top\text{-}k}}
\newcommand{\Sset}{\mathcal S}\newcommand{\Rset}{\mathcal R}
\begin{document}
\twocolumn[{
\begin{center}
\vspace*{2pt}
\rule{0.94\textwidth}{2.5pt}\par
\vspace{18pt}
\noindent\parbox{0.94\textwidth}{\centering
{\fontsize{17}{21.5}\selectfont\bfseries Adaptive DHiSparse:\\[1pt]
\mbox{Speculative Decoding over a Load-Adaptive KV Hierarchy}\par}}\par
\vspace{18pt}
\rule{0.94\textwidth}{1.1pt}\par
\vspace{15pt}
{\large\bfseries Naji Alazhar \qquad Wassel Alazhar\par}
\vspace{5pt}
{\normalsize Umans AI\par}
\vspace{4pt}
{\normalsize\href{mailto:research@umans.ai}{\textcolor{magenta}{\texttt{research@umans.ai}}}\par}
\vspace{8pt}
\end{center}
}]
\begin{center}
{\fontsize{12}{15}\selectfont\bfseries Abstract}
\end{center}
\vspace{2pt}
\noindent
Long-context serving runs out of KV memory before the hardware reaches its useful throughput. HiSparse stores history in host memory and fetches only the selected entries, but it uses one cache budget at every load and does not handle draft-model speculative decoding such as DSpark and DFlash2. We present \method{}, which supports speculative decoding over hierarchical sparse KV and adapts residency to load. It protects the required working set, retains additional history while GPU memory is spare, and recovers the rest through exact fused retrieval. DSpark and DFlash2 keep their proposal and acceptance rules: queries share missing-row retrieval, draft KV uses a window-bounded ring, and verification staging is reused across layer groups. At 100K context and concurrency 128, GLM-5.3 with DFlash2 delivers 793 output tokens/s per decode GPU, 1.45$\times$ HiSparse and 1.27$\times$ non-speculative adaptive serving. Held to the same latency bounds as HiSparse, it sustains up to 4$\times$ the concurrency and 2.3 to 3.7$\times$ the output, depending on the per-token bound. On DeepSeek-V4.1-Flash at 500K context and concurrency 512, Adaptive DHiSparse with DSpark delivers 1.85$\times$ the output of resident serving with DSpark. In a load step, optional residency reduces host fetches from 14.2\% to 1.4\% of selected rows at low load and releases memory under pressure.

\vspace{8pt}
\section{Introduction}
Long-context agents and assistants retain large histories while producing relatively short continuations; a coding request may carry material for several later decisions, leaving the cost of its KV state in place between decoding steps. Sparse attention reduces the entries read by each query, but conventional serving still reserves the complete history in GPU high-bandwidth memory (HBM). As context length and the number of sessions increase, this storage can prevent further admission before the hardware reaches its useful throughput \citep{deepseekv32,deepseekv41,hisparse}. Reducing the computation of a query therefore makes the placement of its history increasingly important.

Hierarchical KV caching addresses this mismatch by retaining history in host memory and recovering only the entries selected by the model. HiSparse combines a bounded GPU cache, fused miss resolution, and prefetch for shared selections \citep{hisparse}. A recently used row can remain in the cache and serve subsequent queries, exploiting locality while bounding each request's device footprint. Its fixed budget imposes a deployment-level trade-off: a small working set permits more sessions but can incur avoidable host reads while HBM is available; a larger reservation reduces retrieval but consumes admission capacity under pressure. A deployment with changing load encounters both sides of this trade-off. Section~\ref{sec:budget} measures it.

HiSparse also has no path for draft-model speculative decoding: it maintains no draft KV, resolves one query per request, and provides no multi-query verification staging. This matters because draft models can amortize target-model computation across several output tokens. DSpark selects request-dependent verification prefixes using confidence and engine cost, while DFlash2 verifies a configured block generated through its conditional selector \citep{dspark,dflash2}. Their benefits depend on both the return from verification and the memory available to sustain concurrent requests. A hierarchy that supports only single-query decoding leaves this acceleration path unavailable.

Verification expands the immediate KV working set because several queries can select the same historical row while retaining separate causal boundaries and producing tentative state that may later be rejected. Resolving each query independently can repeat host reads; retaining every selection independently can require substantially more staging than the hot cache. Draft KV introduces another persistent allocation, and keeping it in a full-history layout can consume the capacity released by target offloading. The KV layer must therefore manage shared historical reads, temporary verification state, and the prefix retained after acceptance as parts of the same execution path.

\begin{figure*}[t]
\centering
\includegraphics[width=\textwidth]{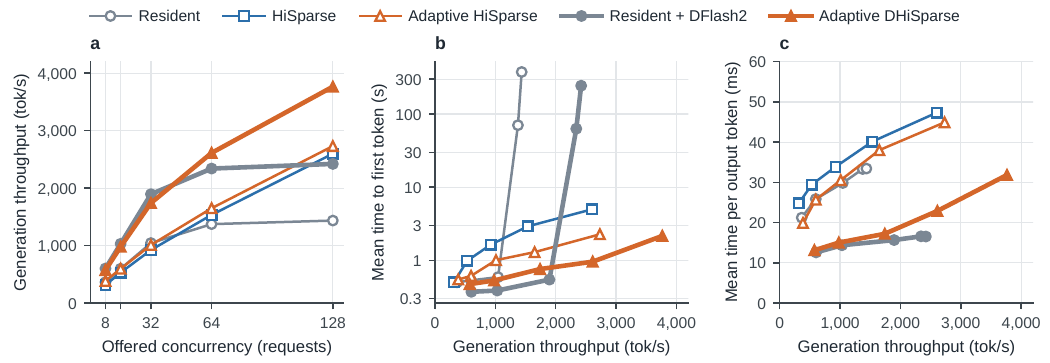}
\caption{\textbf{GLM-5.3 at 32K input and 8K output cap, using the input and output lengths of the HiSparse evaluation.} (a) Generation throughput against offered concurrency, (b) mean TTFT on a logarithmic axis and (c) mean TPOT against generation throughput.}
\label{fig:glm32k}
\end{figure*}

We introduce \method{}, which brings DSpark and DFlash2 to hierarchical sparse KV serving and adapts history residency to the memory available at each load. We call the same system without a drafter Adaptive HiSparse. We protect a fixed working set for exact recovery, retain additional immutable history while HBM is spare, and reclaim that history when other requests need its pages. The fixed hot-cache structure remains small as optional residency grows, allowing the resolver to exploit additional hits without scanning the full resident history. Under pressure, backup and reclamation make those pages available for admission while selected entries remain recoverable from host memory. This couples the allocation decision to the execution that uses it.

We follow the reuse structure of each model: compatible layers in DeepSeek-V4.1-Flash consume a shared global KV source, whereas GLM-5.3 shares sparse indices while retaining distinct layer payloads \citep{deepseekv41,glm53}. Shared positions can consequently announce several future copies without making their bytes interchangeable. We exploit both forms of reuse and bound the additional speculative storage by its lifetime: the drafter keeps its existing visible window in a ring, while target layers reuse staging after their consumers complete. Together, these choices let verification operate over the hierarchy while preserving proposal generation, sparse selections, and acceptance.

We evaluate GLM-5.3 and DeepSeek-V4.1-Flash using matched cache settings, reporting output together with first-token and per-token latency. The experiments cover fixed offered loads, a same-build load step, separate resolver measurements, and a sweep of the hot-cache budget with kernel profiles. They distinguish the capacity of an allocation from the activity that serving actually sustains, and show where more active requests improve aggregate output while slowing individual requests. Comparisons emphasize relative performance under the shared setup; absolute rates are in Table~\ref{tab:headline} and the figures. Our contributions are:
\begin{itemize}
\item \textbf{Speculative serving on hierarchical KV.} At 100K context and offered concurrency 128 (C128), GLM-5.3 with DFlash2 delivers 793 tokens/s per decode GPU, 1.45$\times$ HiSparse and 1.27$\times$ non-speculative adaptive serving, with mean time to first token (TTFT) 2.1$\times$ lower than HiSparse; held to the same latency bounds as HiSparse, it sustains up to 4$\times$ the concurrency and 2.3 to 3.7$\times$ the output, depending on the bound on time per output token (TPOT).
\item \textbf{Residency that follows load.} Optional history reduces host-fetched selections from 14.2\% to 1.4\% at low load in the on/off experiment, yields its memory under pressure, and repopulates as requests drain.
\item \textbf{Bounded speculative state.} A windowed draft ring and reusable verification staging add 0.060 GiB of required GPU state per GLM session while supporting the full configured verification block.
\end{itemize}

Figure~\ref{fig:glm32k} gives an early view at 32K input and an 8K output cap, matching the input/output lengths of the HiSparse evaluation. There, Adaptive DHiSparse delivers 1.7 to 1.9$\times$ the output of HiSparse from C8 to C64 and 1.45$\times$ at C128, where its mean TTFT is 2.3$\times$ lower. Resident serving with DFlash2 is 5 to 9\% ahead up to C32, while every history still fits, and stops scaling at C64. Our main GLM experiments use 100K context, and the DeepSeek experiments use 500K context.

\section{Background}
\subsection{Sparse Attention and KV Residency}
At query position $t$, a sparse indexer assigns a score $I_{t,s}$ to each historical KV entry $s$ and selects $k$ entries, denoted by $\Sset_t$. The model's sparse-attention operator $\operatorname{Attn}$ combines the query hidden state $\mathbf h_t$ with the selected stored rows $\mathbf c_s$ to produce $\mathbf u_t$:
\begin{equation}
 \begin{aligned}
 \Sset_t&=\{s\mid I_{t,s}\in\Topk(I_{t,:})\},\\
 \mathbf u_t&=\operatorname{Attn}\!\left(\mathbf h_t,\{\mathbf c_s:s\in\Sset_t\}\right).
 \end{aligned}
 \label{eq:selection}
\end{equation}
The stored rows may be compressed latent representations \citep{deepseekv32}. We keep $\mathcal{S}_t$ as the ordered list the model's top-$k$ returns, with its tie handling, causal masking and any additional local-attention branch, and write set operations on it when only membership matters. The indexer runs before the selected payload is read, providing an explicit demand signal to the memory system. Keeping selection state on the GPU lets that computation continue over the logical history while the larger attention payload is distributed across memory tiers. The resolver then supplies the announced rows before attention consumes them.

An entry omitted from $\Sset_t$ can appear in a later selection, so sparse attention reduces what a query reads while the complete history remains logically available. HiSparse exploits this distinction by changing where entries are stored rather than which entries attention selects \citep{hisparse}. Let $\Rset_t$ contain the entries ready on the GPU in the tail, hot cache, or optional history. The selected entries that require host recovery form
\begin{equation}
 \mathcal D_t=\Sset_t\setminus\Rset_t.
 \label{eq:demand}
\end{equation}
Placement changes $\mathcal D_t$ while preserving $\Sset_t$. Retaining additional history may reduce recovery, but consumes memory that could admit another request, and populating that history also incurs work.

\subsection{KV Residency Strategies}
Serving approaches differ in how much history they reserve on the GPU and how selected entries reach attention, with the resulting storage and retrieval costs determining the useful active batch.

\paragraph{Full-history residency}
Full-resident serving keeps every admitted request's KV cache in HBM, giving each selected position an existing device location. For a source containing $L$ historical rows, payload storage grows in proportion to $L$. Across concurrent requests, these histories can exhaust the available cache budget and cap the active decode batch \citep{hisparse}. Paging reduces allocation waste, and source sharing removes duplicated state, but each distinct resident history still occupies device memory \citep{pagedattention,deepseekv41}.

\paragraph{Fixed-cache offloading}
HiSparse retains complete history in pinned host memory and gives each request a bounded GPU cache of $B$ entries for each KV source \citep{hisparse}. After the indexer selects positions, a fused kernel identifies hits, protects useful cached rows, chooses least-recently-used (LRU) victims, fetches missing entries, and returns their device locations. Compact indexer state remains on the GPU, allowing the model to select across the full history while its larger KV payload uses the hierarchy.

The cache preserves temporal locality: a selected entry can remain in the hot set rather than being copied on every step. Increasing $B$ retains more history, whereas decreasing it leaves room for additional requests. For models that share selections across layers, HiSparse also prefetches layer-local payload after an earlier index producer announces the positions \citep{hisparse,indexcache}. The transfer can begin during intervening computation and complete before the consuming attention layer.

A fixed cache keeps its configured size at every load: it fetches rows from host while memory is spare, and a larger reservation would cost admission capacity when load rises. This motivates separating the small replacement working set from additional, reclaimable history.

\paragraph{Physical-source sharing}
A KV source produces bytes consumed by one or more layers: different layers may select the same positions from distinct sources, whereas DeepSeek-V4.1-Flash shares global KV and indexer keys across compatible layers. A Full layer produces a source and a selection; a Reindex layer selects again using shared KV; a Reuse layer consumes the latest applicable selection \citep{deepseekv41}. A residency manager can consequently organize payload by request and physical source, while selection reuse determines when a location vector can also be shared.

\subsection{Parallel Drafting and Verification}
Speculative decoding proposes several tokens for verification in one target pass: DFlash conditions a parallel draft block on target features \citep{dflash}; DSpark adds a lightweight sequential head and confidence-scheduled verification prefixes \citep{dspark}; DFlash2 uses a conditional selector over its candidate lattice \citep{dflash2}. The target accepts a prefix and emits a correction or bonus according to the existing verification rule \citep{specsampling}. These methods determine proposal and acceptance behavior; the KV layer supplies the historical and tentative state required to execute their queries.

\section{Architecture}
\label{sec:architecture}
\begin{figure*}[t]
\centering
\includegraphics[width=\textwidth]{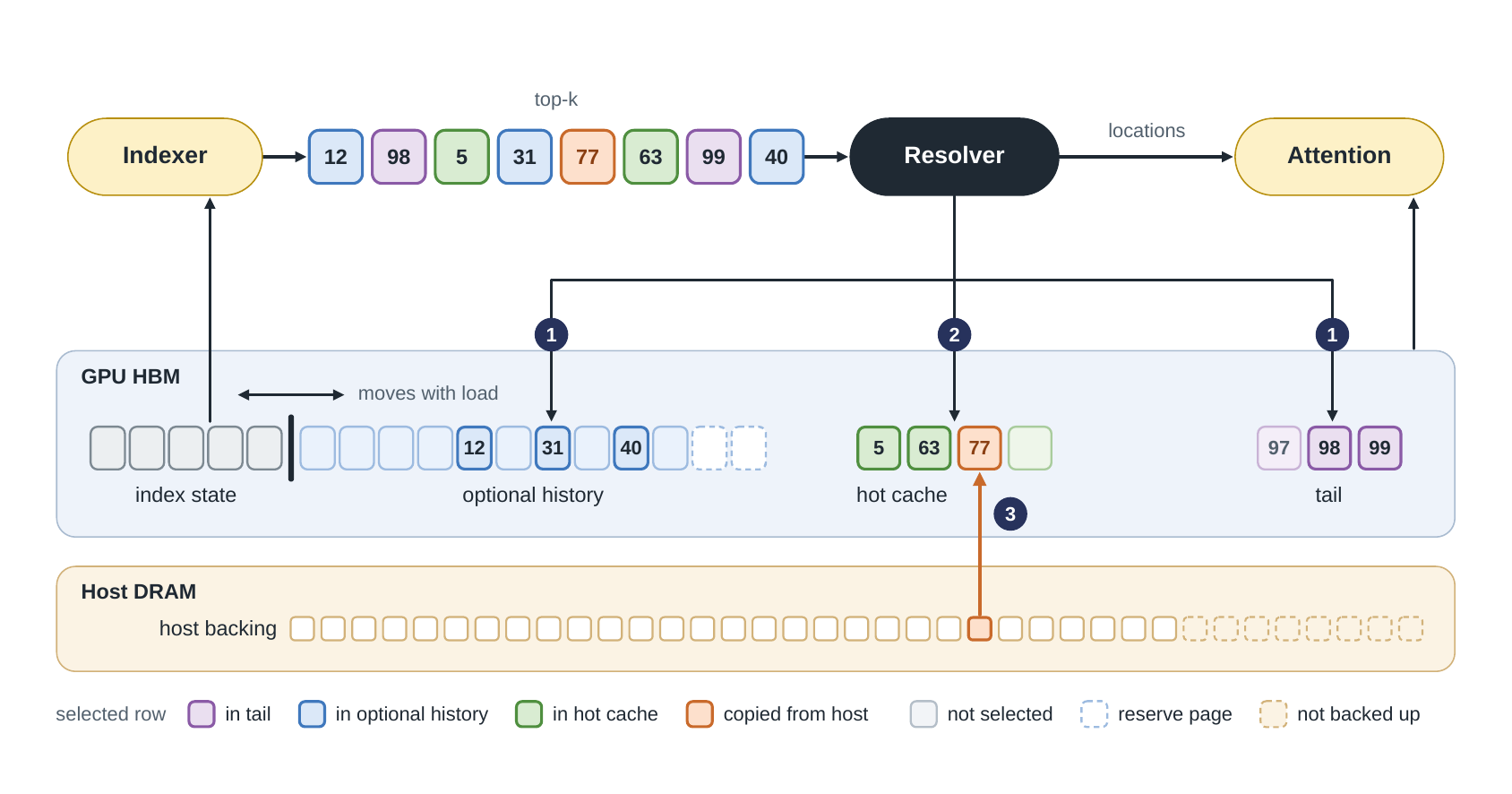}
\caption{\textbf{One decode step and where its KV lives.} The indexer reads index state and emits the top-$k$ positions, colored here by where each row is found. The resolver first takes rows that are in the tail or in optional history (1), then scans the hot cache (2). A row found in none of them is copied from host DRAM into the hot cache (3) before attention reads the selected rows. Optional history holds imported history in reclaimable capacity of the KV pools, drawn schematically as one divider that moves with load. The hot cache and the tail are kept per request and KV source and are never lent. Host backup can be deferred while HBM is spare: a row without host backing stays protected on the GPU, so every missing row has host backing. A lookup skips steps 2 and 3 only when every selected row resolves in step 1. Row positions are illustrative.}
\label{fig:decode}
\end{figure*}

We combine load-adaptive residency, residency-aware resolution, and speculative KV-state management around the selected set in equation~\eqref{eq:selection}: the model chooses positions, and our runtime supplies their device locations. Figure~\ref{fig:decode} follows one decode step from selection to attention. We first describe the division between required storage and optional history, then how resolution and speculative execution use it.

\subsection{Load-Adaptive Residency}
\label{sec:residency}
We reserve a hot cache, a mutable tail and metadata for each admitted request and KV source. With a drafter we also reserve verification staging (Section~\ref{sec:bounded-spec}). The hot cache holds up to $B$ recently useful rows; the tail holds newly computed state until it can become immutable history. Additional history occupies pages that can be returned when required allocations grow, letting a request retain many more than $B$ rows while keeping the replacement structure small.

Optional history occupies reclaimable capacity within the configured KV pools: unclaimed dedicated payload pages and eligible unused index backing. Let $M$ be the capacity of those pools. At time $\tau$, let $M_{\mathrm{req}}(\tau)$ be the memory held by required allocations and $M_{\mathrm{opt}}(\tau)$ the memory held by optional history. Keeping an admission reserve $M_{\mathrm{res}}$, we require
\begin{equation}
M_{\mathrm{req}}(\tau) + M_{\mathrm{opt}}(\tau) + M_{\mathrm{res}} \le M .
\label{eq:budget}
\end{equation}
Required allocations and the reserve take priority; hot-cache and tail storage are never lent.

Optional residency covers complete pages of a request's imported history. Rows generated during decoding and the partial boundary page stay under the tail and hot-cache rules, and a row may be in the optional map and in the hot cache at once. Let $a_r$ be the number of optional pages held by request $r$. When an admission leaves a deficit of $\Delta$ pages, the runtime reclaims whole requests: it walks the admitted requests in its reclaim order $r_1, r_2, \dots$ and takes the pages of the first $q$, with
\begin{equation}
q = \min\Bigl\{\, q' : \sum_{i=1}^{q'} a_{r_i} \ge \Delta \,\Bigr\}.
\label{eq:reclaim}
\end{equation}
The other requests keep their pages. If the optional pages do not cover $\Delta$, the admission waits. When pages are free again, a bounded round-robin scheduler repopulates history, a limited number of pages per attempt.

We track ready locations separately from allocated pages so attention uses rows only after their copies complete. A pending page remains unavailable through the optional map, while the hot-cache recovery path continues to serve selections. The fixed $B$ therefore governs replacement work even when optional history is much larger. Increasing optional residency retains more possible future selections without enlarging the LRU scan, so allocation can respond to available capacity while execution retains the same bounded fallback during admission and partial refill.

We reclaim pages only after their rows have host backing and earlier readers complete, then return them to required allocations. Population reverses this order: reserve eligible pages, copy or retain their bytes, and publish ready locations after completion. This ordering lets a request continue through partial coverage while protecting device-only rows until their host backup completes. Every row of a request's history is therefore, at all times, ready on the GPU or has a host copy.

On admission we reserve the request's hot cache and keep its imported history on device while HBM is spare, deferring its host backup. We call this resident-first admission. Under pressure, we select whole requests until their optional pages cover the deficit, back up their history, and retire those pages once backup and earlier readers complete; the new request takes the freed capacity and the other requests keep their history on device. A reclaimed request then resolves its history through its hot cache, fetching misses from host. Completion releases a request's capacity, which later population refills with history or leaves free.

\subsection{Residency-Aware Resolution}
\label{sec:fused}\label{sec:reuse}
The fused resolver takes the model's selected positions and returns their physical KV locations to attention. It first locates rows in the tail and ready optional-history map (Figure~\ref{fig:decode}, callout~1), then scans the bounded hot cache for unresolved positions, protecting selected rows and identifying unselected LRU victims (callout~2). It recovers the remaining misses from pinned host memory and emits locations in selection order (callout~3). Hit detection, replacement, retrieval, and location generation share request-local intermediate state. The copy path launches over a grid that scales with the batch, from 4 to 32 thread blocks, so that row movement uses the available parallelism.

Rows resolved in the first step (callout~1) never reach replacement or host recovery. In a census of selected rows on GLM-5.3 at C8, without speculation, 69.7\% resolved directly, through the optional-history map or the tail, 30.3\% through the hot cache and 0.05\% from host. The hot scan remains bounded by $B$ rather than the entire resident history. A lookup skips the hot-cache scan only when every selected row resolves through a ready mapping or the tail. Rows generated during decoding follow hot-cache rules, so keeping the whole imported history on device does not by itself make a lookup wholly direct (Section~\ref{sec:resolution}).

We organize the cache by physical KV source so that a DeepSeek-V4.1-Flash Reuse consumer can inherit the preceding selected-location vector together with its materialized payload. A Reindex consumer chooses a fresh top-$k$ from the shared source, reusing rows already present and recovering any new misses. The cache update follows earlier attention readers, allowing several layers to share storage safely as their selections change. A shared source therefore carries one cache state across its consumers: rows fetched for an earlier selection remain available to later layers, and a fresh selection adds only the rows absent from that state. Reusing a selection also avoids another resolution call for the same location vector.

GLM-5.3 instead has 21 index producers announcing positions for 78 distinct payload layers, allowing one selection to start several follower-layer transfers while each layer retains its own bytes. We use HiSparse's exact layer-wise prefetch \citep{hisparse} with source-specific locations, reusing selection and transfer planning through shared indices and reusing payload itself where the model shares KV. This distinction determines which consumers share a resolved vector and which need separate transient storage, whose size we bound by the simultaneously live layer group in Section~\ref{sec:bounded-spec}.

\subsection{Speculative Verification over the Hierarchy}
\label{sec:dspark}
\begin{figure*}[t]
\centering
\includegraphics[width=\textwidth]{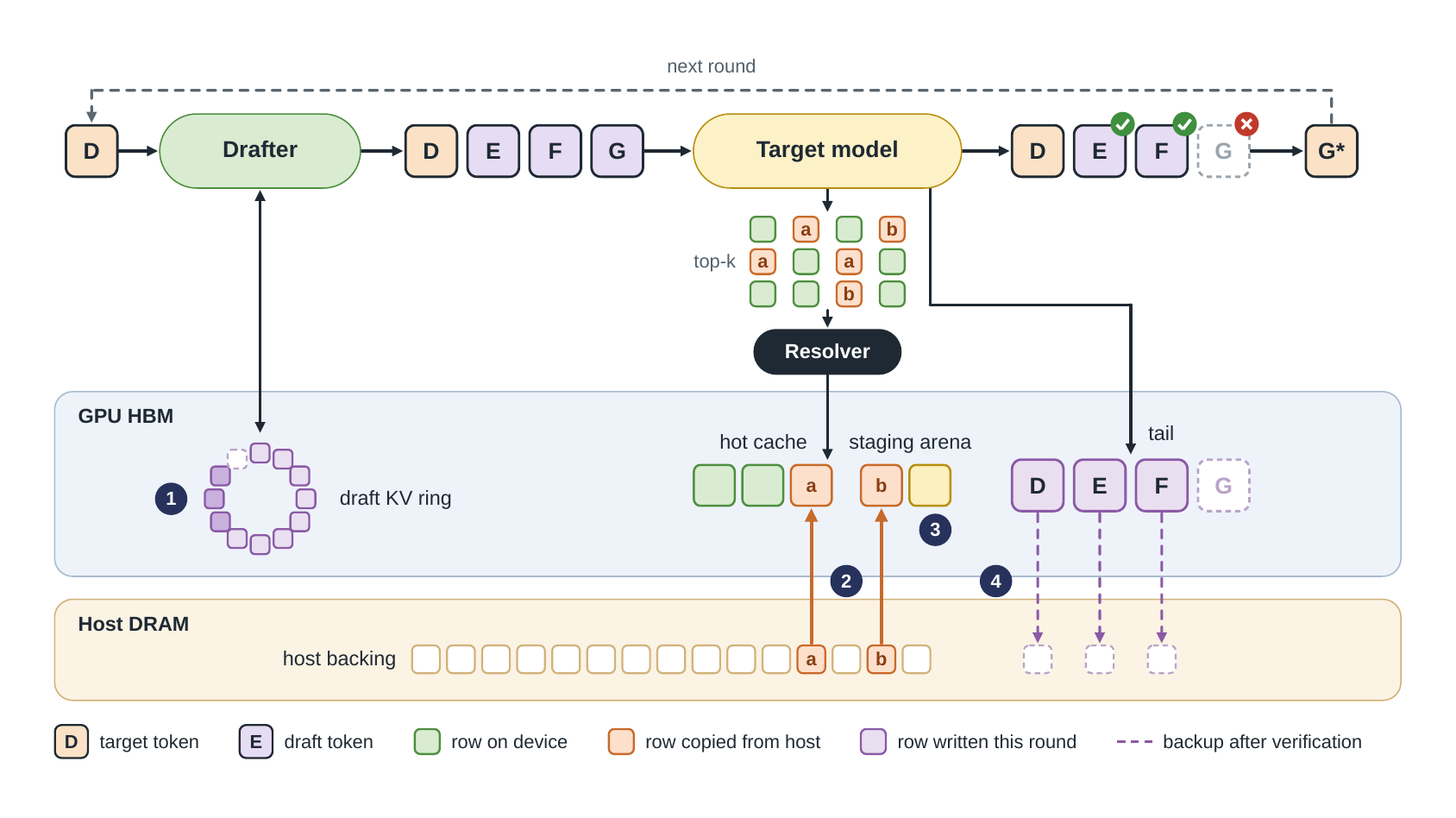}
\caption{\textbf{One speculative round over the KV hierarchy.} The drafter, conditioned on target features, extends the anchor D into a block, and the target model verifies its $K$ queries in one pass. Here E and F are accepted, G is rejected, and the correction G$^{*}$ opens the next round. (1) Draft KV is held in a ring bounded by the drafter's window. (2) For each request and KV source, the resolver handles the top-$k$ selections of the $K$ queries together, so each distinct missing row is copied from host once (a is selected by three queries, b by two), and each query still attends over its own selection. (3) Rows that do not fit the hot cache go to a staging arena, reused across layer groups after its readers finish and promoted rows are copied back. (4) The target writes the rows of all $K$ queries to the tail. Acceptance commits the computed rows of the anchor and the accepted drafts; the rejected row is dropped and G$^{*}$ has no row until the next forward. Host backup of the committed rows follows verification. Sizes are illustrative ($K = 4$, $k = 3$).}
\label{fig:speculative}
\end{figure*}

DSpark chooses a per-request prefix with its confidence scheduler, whereas DFlash2 uses a fixed block \citep{dspark,dflash2}. For request $r$, we denote the number of target queries in a round by $K_r$, including the anchor, and the maximum supported width by $K_{\max}$. Our KV path leaves proposals and acceptance unchanged. Figure~\ref{fig:speculative} shows how the common retrieval and state-management path handles the resulting block.

Let $n_r$ be the number of token positions covered by committed target KV before the round. Query $j$, with $0 \le j < K_r$, sits at position $n_r+j$; $j=0$ is the anchor. Its visible token positions form
\begin{equation}
\mathcal{V}_{r,j} = \{\, v : v < n_r \,\} \cup \{\, n_r + u : 0 \le u \le j \,\},
\label{eq:visibility}
\end{equation}
the committed history, served through the hierarchy, and the current block up to the query itself, read from the tail. We write $\Sset_{r,j}$ for the query's selected rows; at token granularity, $\Sset_{r,j}\subseteq\mathcal V_{r,j}$. Compressed source rows must also be complete, as described below. We keep these query boundaries when resolving historical demand, even though all queries execute in one target pass.

For a fixed request and payload source, let $\mathcal{R}_r$ be the rows ready on the GPU, as in equation~\eqref{eq:demand}, and $\mathcal{H}_r$ the rows with a complete host copy. The retrieval union $\mathcal{U}_r$ and the missing-row set $\mathcal{D}_r$ are
\begin{equation}
\begin{aligned}
\mathcal{U}_r &= \bigcup_{j=0}^{K_r-1} \mathcal{S}_{r,j}, \qquad |\mathcal{U}_r| \le K_r k,\\
\mathcal{D}_r &= \mathcal{U}_r \setminus \mathcal{R}_r \subseteq \mathcal{H}_r .
\end{aligned}
\label{eq:verify-union}
\end{equation}
The inclusion is the invariant of Section~\ref{sec:residency}: a row that is not ready on the GPU always has a host copy.
We deduplicate these misses within the request and payload source, copy each distinct row once, and return its location to every selecting query. In Figure~\ref{fig:speculative}, callout~2, row a is selected three times and row b twice, yet each requires one host copy while attention retains each ordered $\Sset_{r,j}$ and its causal mask. Newly computed rows and incomplete boundary state remain in the tail, where their visibility follows the individual query.

A union can exceed the $B$-row hot cache even when each single-query selection fits. We reserve staging for these overflow rows and protect all selected locations until their attention consumers finish. The bound $K_{\max}k$ per live request/source supplies space for disjoint selections; overlapping selections reduce the copies required within that allocation. Thus the configured hot-cache budget continues to preserve temporal locality while verification can materialize a larger immediate working set. Section~\ref{sec:bounded-spec} describes how layer groups reuse this staging.

Write $x_{r,0}$ for the anchor and $x_{r,1},\dots,x_{r,K_r-1}$ for the proposed tokens. The existing acceptance procedure \citep{specsampling} accepts $A_r$ of the $K_r - 1$ proposals, $0 \le A_r \le K_r - 1$, and emits a correction or bonus $b_r$ that starts the next round. The retained state and new committed frontier $n'_r$ satisfy
\begin{equation}
 \begin{aligned}
 \text{new output}&=[x_{r,1},\ldots,x_{r,A_r},b_r],\\
 \text{committed prefix}&=[x_{r,0},\ldots,x_{r,A_r}],\\
 n'_r&=n_r+1+A_r.
 \end{aligned}
 \label{eq:dspark-state}
\end{equation}
The next forward computes the target KV of $b_r$. In Figure~\ref{fig:speculative}, callout~4, the target has computed D, E, F, and G; accepting E and F retains D--F, discards G, and emits G$^*$ as the next anchor.

For a source that packs $m$ consecutive token positions into each stored row ($m=1$ without compression), the source-row indices $s$ completed by the round satisfy
\begin{equation}
\left\lfloor n_r/m \right\rfloor \le s < \left\lfloor n'_r/m \right\rfloor .
\label{eq:complete-rows}
\end{equation}
These rows join the immutable history, and their host backup is scheduled after verification; any remaining incomplete group stays in the tail until it is complete. Committed rows follow the accepted prefix, rejected and unverified rows are never published, and the drafter's state advances with the same prefix.

This interface separates how many queries the drafter asks the target to verify from how much history those queries fetch. DSpark can change $K_r$ across requests and rounds, while DFlash2 uses its configured width; both retain their original verification decisions. The storage system handles the resulting union and accepted state, including repetitions of a selected position across queries.

\subsection{Bounded Speculative Storage}
\label{sec:bounded-spec}
Speculation adds persistent draft KV and temporary target-verification state, allocations that can become the next concurrency constraint once the target history is offloaded. We size the draft by its visible window and the target staging by the layer payloads that are simultaneously live, preserving the information consumed by the models while limiting its physical footprint.

A drafter that reads at most $W$ prior positions needs storage for that window and its current block of at most $K_{\max}$ positions. With $p$ rows per page, alignment adds at most $p-1$ positions. We allocate a ring of $R_d$ rows per request, where
\begin{equation}
 R_d=p\left\lceil\frac{W+p-1+K_{\max}}{p}\right\rceil.
 \label{eq:ring-size}
\end{equation}
This covers the visible interval when previous readers complete before overwrite. For request slot $i$, the physical ring address $\phi(i,t)$ of draft position $t$ is
\begin{equation}
\phi(i,t)=iR_d + (t \bmod R_d).
\label{eq:ring-address}
\end{equation}
Two live positions of one request differ by less than $R_d$, so they never share a row, and slots occupy disjoint ranges. Figure~\ref{fig:speculative}, callout~1, places this bounded draft state in HBM. Logical positions advance through the ring while the drafter continues to read the same suffix, making its KV allocation independent of the length of the target's full context.

In GLM-5.3/DFlash2, $W=2{,}048$, $p=64$, and $K_{\max}=8$ give $R_d=2{,}176$, a page-rounded size that also covers the six- and four-query blocks. Six draft transformer layers have twelve K/V allocations totaling 12,288 bytes per row. For 35 request slots they occupy about 0.87 GiB per GPU. The target can continue selecting historical rows far outside the draft window through the KV hierarchy, while the drafter retains the suffix and target features it already uses.

Target verification instead reuses temporary storage, with overflow rows in Figure~\ref{fig:speculative}, callout~3, occupying staging separate from the hot cache. Layers that share selected positions may still read different payloads, so each simultaneously active source receives a distinct arena. After all consumers of a layer group complete, the next group can reuse those arenas. Promoted rows are copied back to their stable hot-cache addresses before an arena is released.

With $\chi$ payload sources live at once, $N$ request slots per GPU, and $d_{\mathrm{KV}}$ bytes per stored KV row, the staging reservation is
\begin{equation}
 M_{\mathrm{stage}}=\chi N K_{\max}k\,d_{\mathrm{KV}}.
 \label{eq:staging-size}
\end{equation}
The reservation $M_{\mathrm{stage}}$ covers the maximum verification width, with allocator rounding, maps, and persistent cache state accounted separately, while each round uses locations selected by its live queries.

\begin{table}[t]
\centering\footnotesize\setlength{\tabcolsep}{5pt}
\caption{\textbf{Memory per session for GLM-5.3, in GiB.} Required GPU memory at 32K and 100K context, and the full host backup at 100K. Model weights and shared workspaces are excluded. $\dagger$ excludes optional residency, which only uses HBM that is spare.}
\label{tab:memory}
\begin{tabular}{lrrr}
\toprule
 & \multicolumn{2}{c}{GPU, required} & Host backup \\
\cmidrule(lr){2-3}
Method & 32K & 100K & 100K \\
\midrule
Resident & 1.456 & 4.444 & -- \\
HiSparse & 0.259 & 0.432 & 4.186 \\
Adaptive HiSparse & 0.259$^{\dagger}$ & 0.432$^{\dagger}$ & 4.186 \\
Resident + DFlash2 & 1.481 & 4.469 & -- \\
Adaptive DHiSparse & 0.319$^{\dagger}$ & 0.492$^{\dagger}$ & 4.186 \\
\bottomrule
\end{tabular}
\end{table}

GLM-5.3 processes its index-sharing layer groups sequentially, with at most four payload layers live together. Reusing four arenas instead of dedicating one to each of 78 layers divides this transient term by 19.5, from 24.0 to 1.23 GiB per GPU at 35 slots. The windowed ring and reusable staging together add 0.060 GiB of required GPU state per session in the evaluated layout. Copy-back remains part of execution cost. Table~\ref{tab:memory} gives the per-session memory: at 100K Adaptive HiSparse requires 0.432 GiB of GPU memory against 4.444 GiB for resident serving, 10.3$\times$ less; with DFlash2 the two figures are 0.492 and 4.469 GiB.

\paragraph{Capacity accounting} Speculation also adds fixed memory on each decode GPU, outside the per-session figures of Table~\ref{tab:memory}: the DFlash2 drafter takes 4.9 GiB, the target model 1.3 GiB more, and verification metadata 0.4 GiB. These allocations reduce the memory left for the KV and index pools. The index pool then holds 3.72 million rows per GPU against 5.37 million, enough for 36 complete histories of 100K input and 2,048 output tokens against 52. We configure 35, which gives 140 sessions on four GPUs against 208.

\subsection{Exactness}
\label{sec:exactness}
Our hierarchy changes where KV rows are stored, while each query receives the model's ordered selection, payload and scales, and causal mask. Sharing a retrieval reuses the copy rather than merging attention inputs. For the payload source at hand, $\pi_r(s)$ is the GPU location the resolver returns for row $s$ of request $r$; a row held in two places has one of them chosen. If $s_{r,j,i}$ is the $i$-th entry in query $j$'s ordered selection, the resolver returns a location $\ell_{r,j,i}$ satisfying
\begin{equation}
\ell_{r,j,i}=\pi_r(s_{r,j,i}), \qquad \mathrm{KV}[\ell_{r,j,i}] = \mathbf{c}_{r,\,s_{r,j,i}} .
\label{eq:location-invariant}
\end{equation}
Here $\mathrm{KV}[\ell]$ is the GPU payload at location $\ell$, and $\mathbf c_{r,s}$ is the original KV row $s$ of request $r$. Each query receives its own ordered locations; queries selecting the same row share its location. We preserve target features, proposal probabilities, acceptance, and accepted-prefix updates, and make committed rows recoverable before reusing their device storage. Given equal model state and numerical execution, these properties preserve attention inputs and the verifier's output law \citep{specsampling}. The ring retains the drafter's visible window, and staging is reused only after its readers and required copy-back complete. These are the design conditions for exact placement.

\section{Experiments}
\label{sec:experiments}
We evaluate GLM-5.3 and DeepSeek-V4.1-Flash serving, then examine how optional residency responds to load, how verification width affects output, what a miss costs, and how large the hot cache should be.

\subsection{Experimental Setup}
\label{sec:setup}
\paragraph{Hardware and workloads}
Experiments use $8\times$B300 nodes (269 GiB of HBM per GPU, 4.3 TB of host memory over PCIe). GLM-5.3 uses four GPUs for prefill and four for decode at 100K context, while DeepSeek-V4.1-Flash uses two per role at 500K. We evaluate end-to-end serving with the closed-loop concurrency benchmark used by HiSparse \citep{hisparse}, sweeping offered concurrency $C$ at a fixed input length and output cap. GLM-5.3 runs use 100K input and a 2,048-token output cap; Figure~\ref{fig:glm32k} uses HiSparse's 32K input and 8K output. DeepSeek-V4.1-Flash runs use 500K input and a 2,048-token output cap. The 100K and 500K workloads reuse prepared prefixes, one for GLM-5.3 and 16 for DeepSeek-V4.1-Flash, with independent decode KV per request. We report output throughput per decode GPU, time to first token (TTFT) and time per output token (TPOT).

\paragraph{Configurations}
We compare resident serving, HiSparse, and our adaptive configurations in SGLang \citep{sglang,hisparse}, with DSpark or DFlash2 where supported. For both models, HiSparse and the adaptive configurations use the same model-computation kernels. All adaptive configurations add the KV management of Section~\ref{sec:architecture}; Section~\ref{sec:glm-results} attributes the gain over HiSparse. GLM offloading configurations use FP8 KV, $k=2{,}048$, and $B=4{,}096$, with cache settings matched across methods \citep{hisparseguide}; Section~\ref{sec:budget} sweeps $B$ in this configuration: at C128 output peaks at $B = 2k$ for both systems. DeepSeek uses our optimized HiSparse port as its reference, with $k = 512$ and hot caches of 8,192 rows for the KV source that five layers select from and 4,096 rows for each of the three sources that one layer selects from. Figure~\ref{fig:glm} specifies the DFlash2 block schedule; Figure~\ref{fig:block} shows the block-size comparison.

\paragraph{Measurement}
Output throughput counts returned tokens per second per decode GPU, excluding prefill GPUs; Figure~\ref{fig:glm32k} reports aggregate output across four decode GPUs. Post-wave latency excludes the first $C$ submitted requests.

\begingroup
\setlength{\parskip}{1pt plus 0.5pt minus 0.5pt}
\titlespacing*{\paragraph}{0pt}{4pt plus 1pt minus 1pt}{0.7em}
\subsection{Experimental Results}

\subsubsection{GLM-5.3 at 100K context}
\label{sec:glm-results}
\begin{figure*}[t]
\centering
\includegraphics[width=\textwidth]{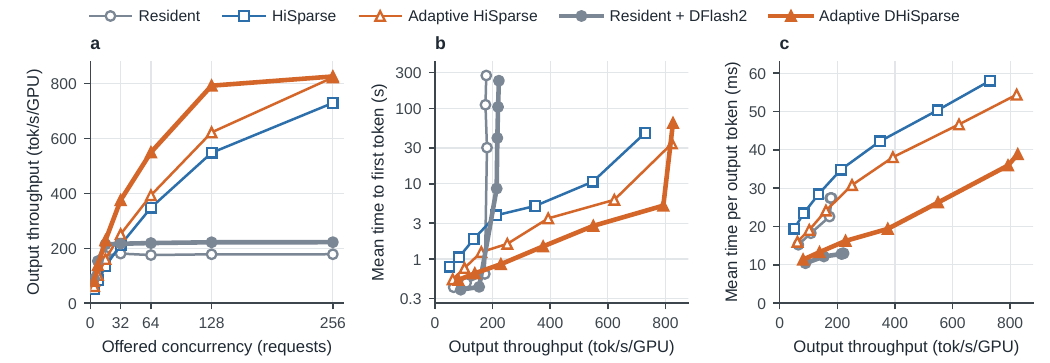}
\caption{\textbf{GLM-5.3 serving at 100K context.} (a) Output throughput against offered concurrency, (b) mean TTFT and (c) mean TPOT against output throughput. Color gives the KV policy and filled markers add DFlash2. TTFT in (b) is on a logarithmic axis. Adaptive DHiSparse uses static blocks of 8 at C4 to C32, 6 at C64 and 4 at C128 and C256.}
\label{fig:glm}
\end{figure*}

Figure~\ref{fig:glm} shows the serving curves and Table~\ref{tab:headline} gives the C128 and C256 results.

\paragraph{Where the 1.45$\times$ comes from} At C128, Adaptive DHiSparse delivers 793 tokens/s per decode GPU, 1.45$\times$ the output of HiSparse. The gain is the product of two steps. The first, from HiSparse to non-speculative adaptive serving, is 1.14$\times$. The copy path launches over a grid that scales with the batch, 4 to 32 thread blocks against a fixed four, exposing more parallelism for row movement. Mean time per token is 7\% lower. Resident-first admission places incoming history directly in GPU storage and defers its host backup until memory pressure, which keeps that backup off the admission path: the prefill/KV handoff falls from 10.4 to 4.0 seconds, for 1.8 seconds of added allocation wait. In a closed loop a shorter wait keeps more of the 128 requests active, 126 against 121. The second step, speculation, is 1.27$\times$. A round verifies four queries and emits 2.88 tokens in 2.2$\times$ the time of a non-speculative step, so each session runs 1.30$\times$ faster while 99\% of the sessions stay active. The gain follows the tokens a round emits: with 2.67 tokens per round instead of 2.88, speculation delivered 1.13 to 1.16$\times$ non-speculative adaptive serving instead of 1.27$\times$, and 1.29 to 1.31$\times$ HiSparse instead of 1.45$\times$. Mean TTFT ends 2.1$\times$ lower than HiSparse and mean TPOT 29\% lower.

\begin{table*}[t]
\centering\footnotesize\setlength{\tabcolsep}{5pt}
\caption{\textbf{GLM-5.3 at 100K context, offered concurrency 128 and 256.} Output is in tokens/s per GPU. Post-wave latency excludes the first $C$ submitted requests. Active is the mean number of requests decoding at once over the central half of a run.}
\label{tab:headline}
\begin{tabular}{llrrrrrrr}
\toprule
 & & & Against & \multicolumn{3}{c}{TTFT (s)} & TPOT & Active \\
\cmidrule(lr){5-7}
$C$ & Method & Output & HiSparse & mean & post-wave & post-wave p99 & (ms) & / slots \\
\midrule
128 & HiSparse & 548 & -- & 10.68 & 5.11 & 26.97 & 50.3 & 121.1 / 208 \\
 & Adaptive HiSparse & 622 & 1.14$\times$ & 6.13 & 2.48 & 11.85 & 46.6 & 125.6 / 208 \\
 & Adaptive DHiSparse & 793 & 1.45$\times$ & 5.13 & 2.25 & 9.67 & 35.9 & 124.7 / 140 \\
\addlinespace
256 & HiSparse & 730 & -- & 47.61 & 42.54 & 102.89 & 58.0 & 190.5 / 208 \\
 & Adaptive HiSparse & 823 & 1.13$\times$ & 34.36 & 31.01 & 81.99 & 54.4 & 201.5 / 208 \\
 & Adaptive DHiSparse & 826 & 1.13$\times$ & 64.62 & 67.79 & 87.60 & 38.9 & 138.2 / 140 \\
\bottomrule
\end{tabular}
\end{table*}

\paragraph{Why speculation stops paying at C256} With $b$ active sessions, output is proportional to $b/\mathrm{TPOT}$, so the output $Q$ of speculative serving ($s$) against non-speculative adaptive serving ($0$) is
\begin{equation}
\frac{Q_s}{Q_0} \approx \frac{b_s}{b_0}\cdot\frac{\mathrm{TPOT}_0}{\mathrm{TPOT}_s} .
\label{eq:spec}
\end{equation}
Speculation pays while it keeps a larger share of the sessions than its TPOT ratio. At C128 it keeps 99\% of the sessions at 0.77 of the TPOT, and the relation gives 1.29$\times$ against 1.27$\times$ measured. At C256 the speculative configuration admits 140 sessions against 208. It keeps 69\% of the sessions at 0.71 of the TPOT, which gives 0.96$\times$ against 1.00$\times$ measured (826 and 823 tokens/s per decode GPU). Speculation then serves fewer sessions, each faster, for the same output, and requests wait longer for a slot: mean TTFT is 1.9$\times$ higher. Both adaptive configurations remain about 13\% above HiSparse.

\paragraph{Why the gap to resident serving grows with load} Resident serving requires 4.444 GiB of GPU memory per session at 100K, against 0.432 GiB for Adaptive HiSparse (Table~\ref{tab:memory}). The memory of 52 offloaded sessions on a GPU therefore holds five resident ones, 20 on four GPUs. With DFlash2 the memory of its 35 sessions holds three, 12 on four GPUs. This is where the resident curves stop scaling: output stays near 178 tokens/s per decode GPU from C32 to C256, and near 220 with DFlash2 from C16, while mean TTFT without DFlash2 grows from 0.6 seconds at C16 to 30 seconds at C32. At C256, Adaptive DHiSparse delivers 4.6$\times$ the output of resident serving and 3.7$\times$ that of resident serving with DFlash2. Below the limit, resident serving has no rows to resolve and stays ahead: at C4 to C16, Adaptive HiSparse is 5 to 7\% below it and 19 to 23\% above HiSparse. A kernel profile at C8 records a shorter step with Adaptive HiSparse (16.6 versus 20.9 ms) and 1.4 ms less cache-kernel work. In the same-build experiment of Section~\ref{sec:loadstep}, optional residency alone makes each session 3.8\% faster. At high load, where optional residency disappears, resident-first admission and the copy path remain active.

\paragraph{Where first-token time goes} Table~\ref{tab:ttft} splits mean TTFT into allocation wait and handoff. At C128, the handoff is 97\% of HiSparse's TTFT. With non-speculative adaptive serving it is 66\%, and with DFlash2 43\%. Below the session limit the handoff is thus the larger part for HiSparse and Adaptive HiSparse, and the adaptive configurations shorten it by 2.6$\times$ and 4.6$\times$. With DFlash2, allocation wait, 49\% of mean TTFT at C128, is already the larger part. At C256 the bottleneck moves to admission: allocation wait is 85 to 97\% of mean TTFT for the adaptive configurations and 67\% for HiSparse. Speculative TTFT stays as high after the first wave of requests, so the pressure is sustained.

\paragraph{Capacity under fixed latency bounds} Table~\ref{tab:latency} lists latency at every load. With post-wave p99 TTFT at most 10 s and mean TPOT at most 40 ms, HiSparse sustains C32 (212 tokens/s per decode GPU), Adaptive HiSparse C64 (393), and Adaptive DHiSparse C128 (793): 4$\times$ the concurrency and 3.7$\times$ the output of HiSparse. The TPOT bound is what holds HiSparse at C32: it reaches 42 ms at C64, while speculation holds 36 ms at C128. With the bound at 50 ms, HiSparse reaches C64 (348) and the output ratio is 2.3$\times$. The ratios also depend on the first-token bound: Adaptive DHiSparse meets it at C128 with 9.67 s, and with a bound of 8 s it sustains C64, 2$\times$ the concurrency and 2.6$\times$ the output of HiSparse.

\subsubsection{DeepSeek-V4.1-Flash at 500K context}
\begin{figure*}[t]
\centering
\includegraphics[width=\textwidth]{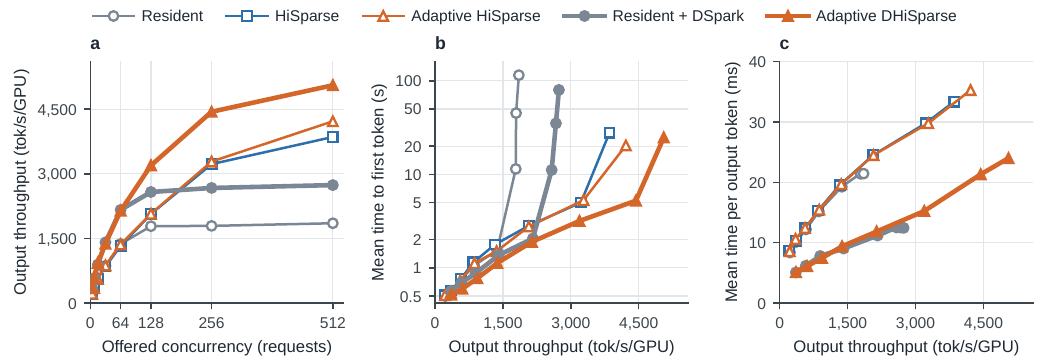}
\caption{\textbf{DeepSeek-V4.1-Flash serving at 500K context.} Panels and encoding as in Figure~\ref{fig:glm}; filled markers add DSpark. HiSparse is our optimized port to DeepSeek-V4.1-Flash.}
\label{fig:deepseek}
\end{figure*}

Figure~\ref{fig:deepseek} repeats the comparison on a second model, at five times the context and with DSpark as the drafter.

\paragraph{The same limit, at a higher load} Resident serving stops scaling at C128: its output stays near 1,800 tokens/s per decode GPU up to C512 while mean TTFT grows from 11 to 115 seconds. Adaptive serving keeps scaling, from 1.16$\times$ resident at C128 to 2.3$\times$ at C512 (4,217 tokens/s per decode GPU), where its mean TTFT is 5.6$\times$ lower. Against HiSparse it is within 4\% up to C256 and 9.4\% higher at C512.

\paragraph{Context length sets the crossover} At C512, adaptive output is 2.6\% below resident at 32,736 tokens, 1.57$\times$ resident at 200K and 2.3$\times$ at 500K. At the shortest context every history fits in HBM and resident serving has nothing to resolve. As the history grows, fewer resident sessions fit and the hierarchy takes over.

\paragraph{Speculation follows the same pattern} With DSpark, adaptive output is 1.85$\times$ resident serving with DSpark at C512 (5,054 tokens/s per decode GPU), with mean TTFT 3.2$\times$ lower. TPOT is 1.9$\times$ higher, because each request shares a step with many more active sessions. The return of speculation over non-speculative adaptive serving falls as load rises, from 1.54$\times$ at C128 to 1.20$\times$ at C512, as on GLM-5.3 above C32.

\subsection{Experimental Analysis}

\subsubsection{What residency buys and what it costs}
\label{sec:loadstep}
\begin{figure*}[t]
\centering
\includegraphics[width=\textwidth]{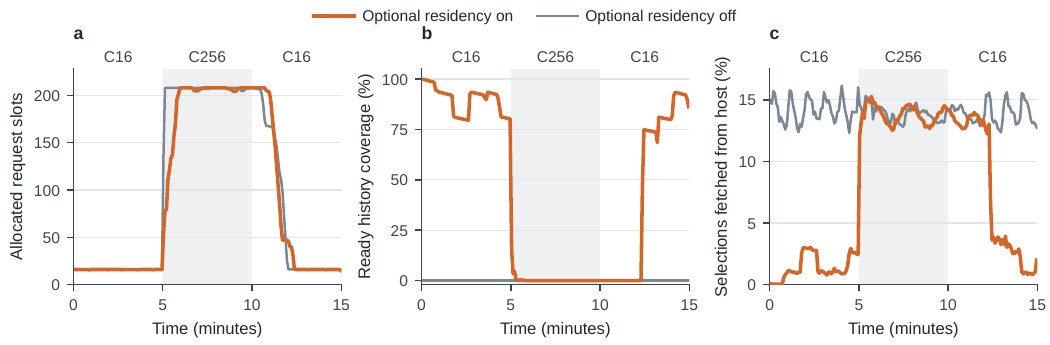}
\caption{\textbf{Residency under a load step.} GLM-5.3 at 100K context, without speculation: the same build with optional residency on and off, offered C16, then C256, then C16 for five minutes each; in-flight requests drain when the load falls. (a) Allocated request slots. (b) Coverage: ready optional history over eligible history. (c) Share of selected rows fetched from host. One-second counters shown in five-second bins. Per-phase request statistics are in Table~\ref{tab:loadstep}.}
\label{fig:loadstep}
\end{figure*}

On GLM-5.3 at 100K context, without speculation, the same build runs with optional residency on and off under a C16--C256--C16 sequence, with five minutes at each offered load (Figure~\ref{fig:loadstep}, Table~\ref{tab:loadstep}).

\paragraph{At low load} With residency on, 75 to 100\% of the eligible history is on the GPU, and the share of selected rows fetched from host is 10.3$\times$ lower. In this first phase per-session speed is 3.8\% higher and time to first token is 1.65 seconds against 3.22.

\paragraph{When load rises} The pages have to be returned before new requests can use them. The runtime backs up and retires the resident history: coverage falls below 1\% within 16 seconds, and the 208 slots fill after 53 seconds, against 3 seconds with residency off. In the C256 phase, allocation wait is 1.4$\times$ longer and per-session speed 3\% lower, with the same host-fetched share in both arms. The prefill/KV handoff takes 8.5 seconds against 26.8, so time to first token in the phase is still 21\% lower, and 11\% more requests complete.

\paragraph{When load falls} In-flight requests drain first, then population refills the freed pages. Coverage is back above 50\% after 142 seconds. The host-fetched share is again 5.8$\times$ lower and per-session speed 13.1\% higher, while time to first token is the same in both arms (0.84 and 0.87 seconds).

While memory is spare, optional residency thus removes 83 to 90\% of host fetches, and sessions run 3.8\% faster before the step and 13.1\% faster after recovery. It costs 3\% of per-session speed and slower slot filling for the minute that follows a load step.

\subsubsection{What a wider verification block returns}
\label{sec:verification-return}
\begin{figure*}[t]
\centering
\includegraphics[width=\textwidth]{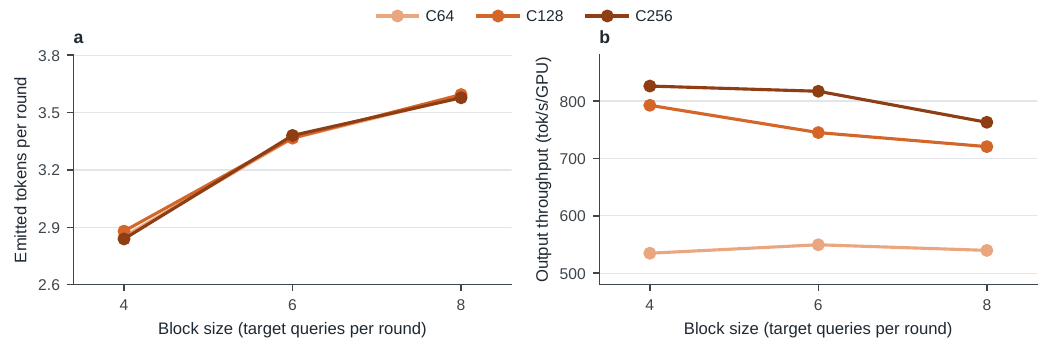}
\caption{\textbf{Block size for DFlash2 at 100K context.} (a) Emitted tokens per speculative round and (b) output throughput for static blocks of 4, 6 and 8 target queries at three offered loads.}
\label{fig:block}
\end{figure*}

Figure~\ref{fig:block} compares static DFlash2 blocks of 4, 6 and 8 target queries at C64, C128 and C256. A wider block emits more tokens per round and takes longer per round. At C64 the two cancel: block 6 emits 18\% more tokens than block 4 in a round that is 17\% longer (Figure~\ref{fig:costbreakdown}a), and the three blocks are within 3\% in output. At higher load the wider blocks delivered less. At C128, blocks of 6 and 8 emit 17\% and 25\% more tokens per round than block 4 but deliver 6\% and 9\% less output, and at C256 they deliver 1\% and 8\% less. At these three loads the serving curve uses the block with the highest output, 6 at C64 and 4 at C128 and C256.

Resolution and copying run alone for 17\% of the block-4 round and 20\% of the block-6 round. Shared retrieval keeps that share down: at C128 with four queries, verification counters show 1.41 missing selections per distinct host-fetched row, so fetching each row once removes 29\% of the fetches the queries would otherwise issue.

\subsubsection{What a miss costs}
\label{sec:resolution}
\begin{figure*}[t]
\centering
\includegraphics[width=\textwidth]{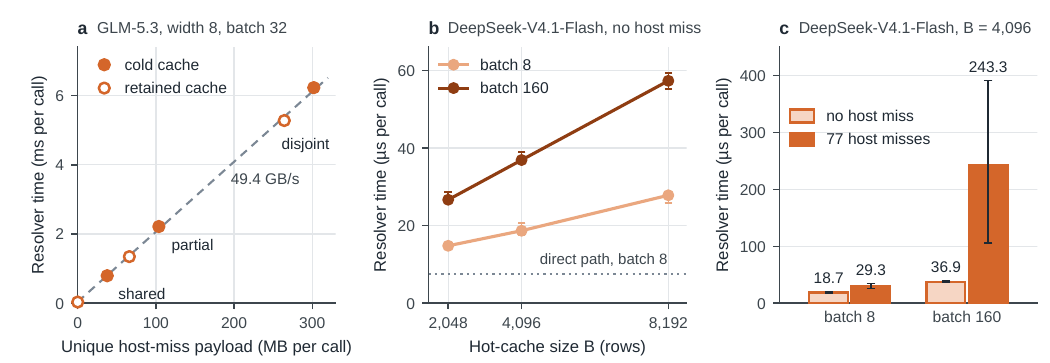}
\caption{\textbf{What resolution costs.} (a) GLM-5.3, verification width 8, batch 32 per GPU, $B = 4{,}096$: resolver time against the unique host-miss payload of a call, for synthetic shared, partial and disjoint selections across the eight queries, with a cold or a retained cache. The dashed line is the least-squares fit, 49.4 GB/s ($R^2 = 0.9992$). One query's top-$k$ over the 32 requests of a call is 37.75 MB, so disjoint selections fetch eight such payloads and shared selections one. (b) DeepSeek-V4.1-Flash at 500K, single-query resolver with no host miss: time against hot-cache size for two batch sizes. The dotted line is the direct path at batch 8. (c) The same resolver at $B = 4{,}096$ with and without 77 host misses per request. Error bars span the minimum and maximum of nine timings in (a) and the 10th to 90th percentiles in (b) and (c). These are isolated kernel measurements with synthetic selections and real payload copies.}
\label{fig:resolver}
\end{figure*}

Figure~\ref{fig:resolver} times the resolver apart from the model.

\paragraph{Time follows bytes} On GLM-5.3 (panel a), at verification width 8, batch 32 per GPU and $B = 4{,}096$, resolver time is linear in the unique host-miss payload of a call. Six cases, from 0.031 ms with no miss to 6.23 ms for 302 MB, fit a rate of 49.4 GB/s ($R^2 = 0.9992$). Eight disjoint selections cost 7.9$\times$ eight identical ones. On GLM-5.3 the cost of resolution is the bytes it moves, and the two mechanisms that remove fetches act on it directly: optional residency, with 83 to 90\% fewer host-fetched rows at low load, and shared retrieval, with 29\% fewer during verification.

\paragraph{The scan grows with the hot cache} On DeepSeek-V4.1-Flash (panel b), a hot cache four times larger makes a call 1.9$\times$ slower at batch 8 and 2.2$\times$ slower at batch 160 with no miss at all, and at $B = 4{,}096$ the direct path is 2.5$\times$ faster than the scan. Optional history adds rows without lengthening the scan. The scan is a small term in serving: on DeepSeek-V4.1-Flash at C128 the whole resolver takes 0.5 to 0.7 ms of a 25 ms step and does not grow with the budget, and on GLM-5.3 at C8 resolver work grows by 0.5 ms per step from $B = k$ to $8k$ (Section~\ref{sec:budget}). Misses add in proportion to the batch (panel c): 77 misses per request add a median of 1.3 $\mu$s per request at both batch sizes, which makes a call 1.6$\times$ slower at batch 8 and 6.6$\times$ slower at batch 160, because the scan grows far less than the batch.

\paragraph{Share of a step} On DeepSeek-V4.1-Flash, resolver kernel work is 0.4 to 1.4\% of the step interval at C16 and 4.75\% at C512 (Figure~\ref{fig:costbreakdown}b). In a GLM-5.3 round with DFlash2 at C64, resolution and copying are the only kernels running for 17 to 20\% of the time (Figure~\ref{fig:costbreakdown}a). Figure~\ref{fig:cachestep} compares the two models at one load, without speculation. In the census of Section~\ref{sec:fused}, about 70\% of selected rows resolved directly and 30\% through the hot cache. Rows generated during decoding go through the hot cache, so no lookup was wholly direct.

\subsubsection{How large the hot cache should be}
\label{sec:budget}
\begin{figure*}[t]
\centering
\includegraphics[width=\textwidth]{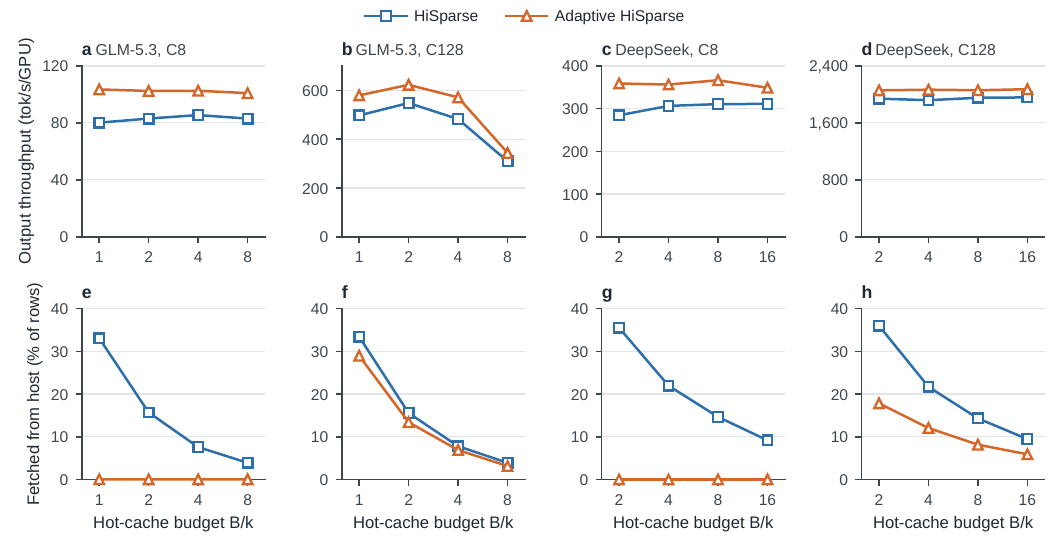}
\caption{\textbf{Output and host fetches against the hot-cache budget.} (a--d) Output per decode GPU and (e--h) share of selected rows fetched from host, for HiSparse and Adaptive HiSparse as $B/k$ varies, without speculation. GLM-5.3 runs at 100K context on one prepared prefix, and DeepSeek-V4.1-Flash at 500K on 16 prepared prefixes.}
\label{fig:budget}
\end{figure*}

We sweep the hot-cache budget from $B = k$ to $8k$ on GLM-5.3 and from $2k$ to $16k$ on DeepSeek-V4.1-Flash, at C8 and C128, without speculation (Figure~\ref{fig:budget}).

\paragraph{On GLM-5.3 the budget trades step time against sessions} Here the sweep changes only $B$: the KV pools keep the size configured for $B = 2k$ and every other setting is as in Section~\ref{sec:setup}, so a larger hot cache leaves room for fewer sessions. At C128, HiSparse's host-miss rate falls from 33\% at $B = k$ to 3.9\% at $B = 8k$, and its TPOT from 57 to 38 ms. Output nevertheless peaks at $B = 2k$ and then collapses: from $2k$ to $8k$ it loses 44\%, and mean TTFT grows from 11 to 115 seconds. Adaptive HiSparse follows the same curve, 14\% higher at the peak, and its profile shows the trade: from $2k$ to $8k$ a decode step shortens from 44.7 to 35.3 ms but holds 13 sessions per GPU instead of 28. A lower miss rate does not buy throughput once the cache takes the memory that sessions need. In these pools, $B = 2k$, the setting of our GLM-5.3 comparisons, gives the highest output of the four budgets for both systems. At C8 the picture changes. With optional residency the host-miss rate is 0.05\% at every budget and output stays within 3\%, so the smallest cache is enough. It is also the fastest: with the miss rate unchanged, resolver work per step grows from 0.35 to 0.85 ms between $B = k$ and $8k$, and TPOT from 18.9 to 19.4 ms. Best against best, Adaptive HiSparse stays 21\% ahead of HiSparse, the gap analyzed in Section~\ref{sec:glm-results}.

\paragraph{On DeepSeek-V4.1-Flash the budget does not matter} At C128, output stays within 2\% across an eightfold range of $B$ for both systems, while the host-miss rate falls from 36\% to 9.5\% for HiSparse and from 18\% to 5.9\% for Adaptive HiSparse. Mean TPOT and mean TTFT are flat as well, the latter at 2.5 to 2.8 seconds. At C8, Adaptive HiSparse fetches nothing from host at any budget, and TPOT shows no trend with the budget for either system; the lower HiSparse output at $B = 2k$ comes from first-token time.

\begin{table}[t]
\centering\footnotesize\setlength{\tabcolsep}{3pt}
\caption{\textbf{Retrieval volume of the two models.} Per generated token and request. Ranges run from the smallest to the largest budget of the sweep, for HiSparse at C128. The hot cache is that of Adaptive HiSparse at the budgets of Section~\ref{sec:setup}.}
\label{tab:budget}
\begin{tabular}{@{}lrr@{}}
\toprule
 & GLM-5.3 & DeepSeek \\
\midrule
Selecting layers $\times$ $k$ & 21 $\times$ 2,048 & 8 $\times$ 512 \\
Selections per token & 43,008 & 4,096 \\
KV sources & 78, one per layer & 4, shared \\
Bytes per missed selection & 2.4 to 2.5 KB & 288 B \\
Host bytes per token & 34 to 4.2 MB & 0.42 to 0.11 MB \\
Hot cache per session & 184 MB & 6 MB \\
Resolver calls per step & 21 & 8 \\
Follower copies per step & 57 & none \\
\bottomrule
\end{tabular}
\end{table}

\paragraph{Why the two models differ} Table~\ref{tab:budget} gives the reason. For each generated token, GLM-5.3 makes 43,008 selections through 21 index producers, and each selection feeds payload layers that keep their own bytes. DeepSeek-V4.1-Flash makes 4,096 selections through eight selecting layers that read four KV sources shared across its layers. A missed selection moves 288 bytes on DeepSeek-V4.1-Flash and about 2.4 KB on GLM-5.3, where it carries the rows of 3.7 payload layers on average. At the smallest budget, where both models miss about a third of their selections, HiSparse fetches 34 MB per generated token on GLM-5.3 and 0.42 MB on DeepSeek-V4.1-Flash, about 80$\times$ less. The DeepSeek-V4.1-Flash hot cache is also about 6 MB per session at the budgets of Section~\ref{sec:setup}, against 184 MB, so enlarging it costs no sessions.

\begin{figure*}[t]
\centering
\includegraphics[width=\textwidth]{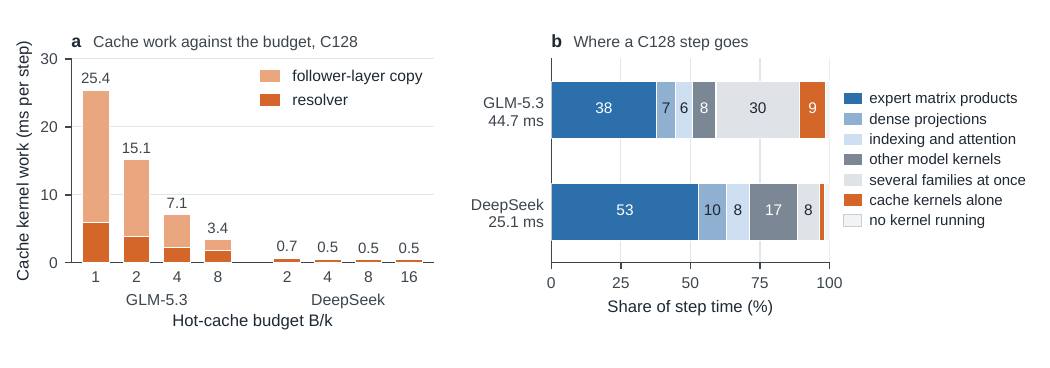}
\caption{\textbf{What the cache costs inside a step, Adaptive HiSparse at C128.} (a) Summed duration of the cache kernels per GPU step as $B/k$ varies, on one scale for both models: the fused resolver and, on GLM-5.3, the copies that load selected rows into the follower payload layers. A GLM-5.3 step holds 24 to 28 sessions per GPU up to $B/k = 4$ and 13 at $B/k = 8$; a DeepSeek-V4.1-Flash step holds 62 to 63. (b) Share of step time by kernel family at the budget used in the paper ($B/k = 2$ for GLM-5.3, 16 for DeepSeek-V4.1-Flash), each family counted where it runs alone. ``Several families at once'' is time in which kernels of more than one family run; ``no kernel running'' is time between kernels, 1.5\% and 1.9\% of the step.}
\label{fig:cachestep}
\end{figure*}

\paragraph{What the cache costs inside a step} Figure~\ref{fig:cachestep} profiles the kernels of Adaptive HiSparse at C128. On GLM-5.3 a step makes 21 resolver calls and 57 follower-layer copies. At $B = 2k$ they take 15.1 ms of kernel work, three quarters of it in the follower copies, and both shrink as the budget grows (panel a); per session, the work falls from 0.55 ms at $2k$ to 0.29 ms at $4k$ and 0.27 ms at $8k$. Most of this work runs under the model's own kernels: 11.0 of the 15.1 ms overlap them, and the cache runs alone for 9\% of the step (panel b). On DeepSeek-V4.1-Flash a step makes eight resolver calls and no follower copy. The resolver takes 0.5 to 0.7 ms of a 25 ms step at every budget, and the residency machinery itself (host backup and page bookkeeping) adds 0.05 ms. Expert matrix products run alone for 38\% of the GLM-5.3 step and 53\% of the DeepSeek-V4.1-Flash step.
\endgroup

\begingroup
\setlength{\parskip}{1pt plus 0.5pt minus 0.5pt}
\titlespacing*{\paragraph}{0pt}{4pt plus 1pt minus 1pt}{0.7em}
\section{Discussion}
\label{sec:discussion}
\label{sec:deployment}

\paragraph{When resident serving is enough} If every active history fits in HBM, resident serving has nothing to resolve: Adaptive HiSparse is 5 to 7\% below it at C4 to C16 on GLM-5.3, and 2.6\% below it on DeepSeek-V4.1-Flash at 32K. The hierarchy pays once resident histories stop fitting, at about 20 sessions for 100K on our GPUs and 12 with DFlash2 (Table~\ref{tab:memory}), and its advantage then grows with both load and context length.

\paragraph{How to operate speculation} Equation~\eqref{eq:spec} says when speculation pays: while the share of sessions it keeps exceeds its TPOT ratio, 0.77 at C128 and 0.71 at C256 in our runs. Once both configurations fill their slots, the share kept is the slot ratio, $140/208 = 0.67$, below the 0.71 measured at C256, so speculation adds no output there and lengthens the wait for a slot. Our schedule narrows the block as load approaches that limit (8 up to C32, 6 at C64, 4 from C128); selecting the block online from load and memory is future work.

\paragraph{What limits capacity next} With KV offloaded, GPU memory is still the binding resource. A session requires 0.432 GiB at 100K, or 0.492 GiB with DFlash2, and speculation adds 6.6 GiB of fixed allocations per GPU for the drafter, the target model and metadata; the configurations admit 208 and 140 sessions. Host memory is far from binding: 208 sessions back up 0.93 TB of a 4.3 TB host, which would hold about 950. At C256, allocation wait is 85 to 97\% of time to first token, and after a load step slots take 53 seconds to fill. The next gains in capacity will come from a lower required state per session and faster reclaim.

\paragraph{Implications for model design} The cost of offloading follows the bytes a token's selections can pull from host. Against GLM-5.3, DeepSeek-V4.1-Flash has a $k$ four times smaller, 2.6$\times$ fewer selecting layers and rows about half as wide, and a selection reads one source row where GLM-5.3 reads 3.7 layer rows. Together these give about 80$\times$ fewer host bytes per token at a comparable miss rate. A DeepSeek-V4.1-Flash step needs eight resolver calls and no follower copy, and its output is insensitive to the cache budget. GLM-5.3 keeps 78 payload layers with their own bytes: a step needs 57 follower copies, and output falls 44\% when the budget grows from $2k$ to $8k$ in pools sized for $2k$ (Section~\ref{sec:budget}). A drafter with a bounded window makes draft KV independent of context: the ring adds 0.025 GiB per session at 32K and at 100K alike.

\paragraph{Scope} Our evaluation covers one GPU platform, two sparse-attention models, and closed-loop workloads at fixed input lengths over a fixed set of prepared prefixes, with cache settings matched across methods. Speculative gains depend on acceptance, which varied between runs of the same workload.
\endgroup

\section{Related Work}
\paragraph{Hierarchical KV caching}
HiSparse establishes LRU-managed hierarchical storage, fused resolution with GPU-assisted host reads, and exact layer-wise prefetch from shared indices \citep{hisparse}, keeping complete history available through a fixed device cache. HiSparse has no draft-model speculative path; we extend it with load-adaptive residency, shared verification retrieval, bounded draft KV, and reusable staging. Strata provides complementary hierarchical context-loading and GPU-assisted I/O mechanisms \citep{strata}.

InfiniGen and ShadowKV use prediction or reconstruction for offloaded execution, and DualDecoder predicts future retrieval \citep{infinigen,shadowkv,dualdecoder}.

\paragraph{Model-side reuse}
Cross-Layer Attention and YOCO share KV state, while IndexCache reuses sparse selections \citep{crosslayer,yoco,indexcache}. DeepSeek-V4.1-Flash combines global-KV and index reuse \citep{deepseekv41}; GLM-5.3 illustrates shared indices with distinct payloads \citep{glm53}. We organize storage around those model-defined relationships, sharing recovered bytes where the model shares a source and preserving distinct bytes where it shares only positions.

\paragraph{Drafting and verification}
Speculative decoding has moved from separate draft models \citep{specsampling} to drafters that reuse the target's heads or features, as in Medusa \citep{cai2024medusa} and the EAGLE series \citep{li2024eagle,li2024eagle2,li2025eagle3}. DSpark develops semi-autoregressive generation on a parallel DFlash backbone and schedules verification prefixes using confidence and engine cost \citep{dflash,dspark}. DFlash2 retains parallel drafting with a conditional candidate-lattice selector \citep{dflash2}; we support both through their existing proposals and acceptance procedures by supplying the KV retrieval and storage required for hierarchical verification.

Two systems study speculative decoding at long context. MagicDec \citep{sadhukhan2025magicdec} shows that speculation can raise throughput at large batch sizes once sequences are long, using drafters with a sparse KV cache. TriForce \citep{sun2024triforce} combines retrieval-based drafting with hierarchical speculation to relieve the KV bottleneck of long sequences, including an offloading setting. Both give the drafter a sparse or partial KV cache; we keep the target's sparse selections exact and manage where its full history lives.

\paragraph{Serving infrastructure}
PagedAttention, SGLang, and DistServe address paged allocation, efficient execution, and prefill--decode separation \citep{pagedattention,sglang,distserve}. Active-request residency complements prefix caching: a prefix hit avoids repeated prefill computation, while the decode hierarchy manages the physical history used by ongoing requests. Our bounded draft and staging allocations operate alongside these serving mechanisms.

\section{Conclusion}
Hierarchical KV caching lets long histories live in host memory, but with a fixed cache budget requests keep fetching from host while HBM sits idle below peak load, and with single-query retrieval draft-model speculation stays outside the hierarchy. Adaptive DHiSparse closes both gaps by bringing draft-model speculative decoding to hierarchical sparse KV serving and adapting history residency to load. A protected working set supports exact recovery, spare HBM retains additional history, and a source-aware resolver shares missing-row retrieval across verification queries. Window-bounded draft KV and reusable staging control the extra memory required by DSpark and DFlash2 while preserving their proposals and accepted-state updates. At 100K context and C128, GLM-5.3 with DFlash2 delivers 1.45$\times$ the output of HiSparse and, held to the same latency bounds, sustains up to 4$\times$ the concurrency. At 500K context and C512, DeepSeek-V4.1-Flash with DSpark delivers 1.85$\times$ the output of resident serving with DSpark.

With Adaptive DHiSparse, GPU residency becomes a runtime decision where it was a deployment constant. Optional history removes most host fetches while HBM is spare and returns its pages when sessions need them. More broadly, sparse attention, hierarchical KV placement and draft-model speculation compose. Each removes a different cost of long-context serving (what a query computes, where its history lives, how many target passes a token takes), and combining them leaves selections, proposals and acceptance unchanged. How cheaply they combine is decided in the model and the drafter. A model that selects few, narrow rows from KV sources shared across its layers leaves the hierarchy little to move (about 80$\times$ fewer host bytes per token on DeepSeek-V4.1-Flash than on GLM-5.3 at a comparable miss rate), and a drafter with a bounded window leaves speculation little to store at any context length.

\FloatBarrier
\bibliographystyle{plainnat}
\begingroup\fontsize{8.5}{10}\selectfont\raggedright\setlength{\parskip}{0pt}\setlength{\bibsep}{0pt}\bibliography{references}\endgroup
\appendix
\begingroup\small
\captionsetup{hypcap=false}
\setcounter{table}{0}
\renewcommand{\thetable}{A\arabic{table}}
\renewcommand{\theHtable}{A.\arabic{table}}
\section{Additional Measurements}
\label{app:additional}
This appendix gives the measurements behind the main results. Table~\ref{tab:loadstep} lists load-step statistics; Table~\ref{tab:ttft} partitions first-token latency into adjacent request intervals; and Table~\ref{tab:latency} gives latency at every measured load. Figure~\ref{fig:costbreakdown} supplies the round and step breakdowns discussed in Sections~\ref{sec:verification-return} and~\ref{sec:resolution}.
\clearpage\onecolumn
\begingroup
\par\vspace{3pt}
\noindent\begin{minipage}{\textwidth}
\centering
\includegraphics[width=0.85\textwidth]{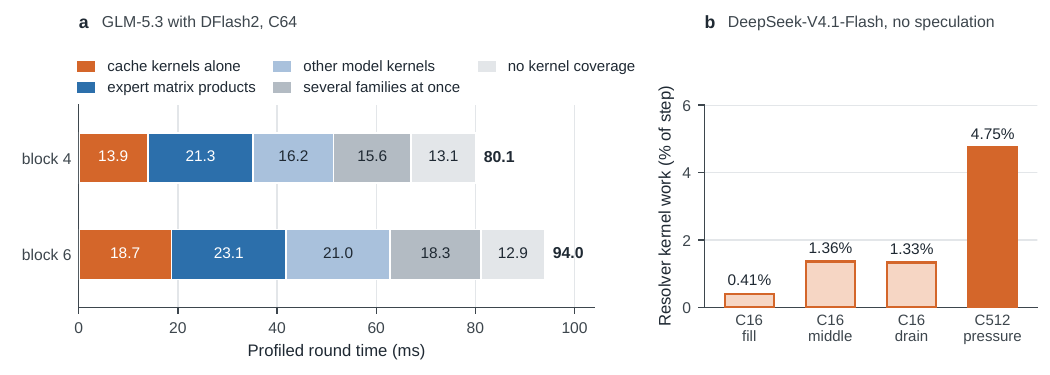}
\captionof{figure}{\textbf{Where step time goes.} (a) GLM-5.3 with DFlash2 at C64: mean profiled round time for static blocks of 4 and 6, as an exclusive partition with a separate category for overlapping kernels. (b) DeepSeek-V4.1-Flash without speculation: resolver kernel work as a share of the step interval in serving profiles, at C16 (fill, middle, drain) and C512 under pressure.}
\label{fig:costbreakdown}
\end{minipage}\par
\endgroup
\vspace{2pt}

\begingroup
\par\vspace{3pt}
\noindent\begin{minipage}{\textwidth}
\centering\footnotesize\setlength{\tabcolsep}{4.2pt}
\captionof{table}{\textbf{Load step, optional residency on / off.} GLM-5.3 at 100K context, without speculation. Each cell gives the value with residency on, then off. The host-fetched share is counted over each phase; the other columns of the first two rows cover requests that start and finish inside the phase. The last row covers the time and the requests after the in-flight count returned to 16.}
\label{tab:loadstep}
\begin{tabular}{lrrrrrrr}
\toprule
Phase & Requests & From host (\%) & TPOT (ms) & Tok/s per session & Gain & TTFT (s) & Alloc.\ wait (s) \\
\midrule
C16 before & 80 / 80 & 1.38 / 14.21 & 25.18 / 26.07 & 39.83 / 38.38 & +3.8\% & 1.65 / 3.22 & 0.03 / 0.00 \\
C256 & 417 / 375 & 13.79 / 13.79 & 53.77 / 52.24 & 18.61 / 19.18 & $-$3.0\% & 36.89 / 46.63 & 28.19 / 19.74 \\
C16 after recovery & 62 / 52 & 2.43 / 13.98 & 24.44 / 28.10 & 40.92 / 36.19 & +13.1\% & 0.84 / 0.87 & 0.02 / 0.01 \\
\bottomrule
\end{tabular}
\end{minipage}\par
\endgroup
\vspace{2pt}

\begingroup
\par\vspace{3pt}
\noindent\begin{minipage}{\textwidth}
\centering\footnotesize\setlength{\tabcolsep}{5pt}
\captionof{table}{\textbf{Where first-token latency is spent, in seconds.} Allocation wait precedes slot allocation; handoff covers prefill readiness and KV transport. The right half excludes the first $C$ submitted requests.}
\label{tab:ttft}
\begin{tabular}{llrrrrrrr}
\toprule
 & & \multicolumn{4}{c}{All requests} & \multicolumn{3}{c}{After the first wave} \\
\cmidrule(lr){3-6}\cmidrule(lr){7-9}
$C$ & Method & Allocation & Handoff & Other & TTFT & Allocation & Handoff & TTFT \\
\midrule
128 & HiSparse & 0.00 & 10.36 & 0.31 & 10.68 & 0.00 & 4.98 & 5.11 \\
 & Adaptive HiSparse & 1.75 & 4.03 & 0.34 & 6.13 & 0.09 & 2.24 & 2.48 \\
 & Adaptive DHiSparse & 2.51 & 2.23 & 0.39 & 5.13 & 0.25 & 1.77 & 2.25 \\
\addlinespace
256 & HiSparse & 32.10 & 14.97 & 0.53 & 47.61 & 33.70 & 8.71 & 42.54 \\
 & Adaptive HiSparse & 29.15 & 4.70 & 0.51 & 34.36 & 27.09 & 3.77 & 31.01 \\
 & Adaptive DHiSparse & 62.50 & 1.50 & 0.61 & 64.62 & 66.46 & 1.09 & 67.79 \\
\bottomrule
\end{tabular}
\end{minipage}\par
\endgroup
\vspace{2pt}

\begingroup
\par\vspace{3pt}
\noindent\begin{minipage}{\textwidth}
\centering\footnotesize\setlength{\tabcolsep}{5pt}
\captionof{table}{\textbf{First-token and per-token latency at every load, GLM-5.3 at 100K.} The middle columns exclude the first $C$ submitted requests.}
\label{tab:latency}
\begin{tabular}{llrrrrrr}
\toprule
 & & \multicolumn{2}{c}{TTFT (s)} & \multicolumn{2}{c}{TTFT after the first wave (s)} & \multicolumn{2}{c}{TPOT (ms)} \\
\cmidrule(lr){3-4}\cmidrule(lr){5-6}\cmidrule(lr){7-8}
Method & $C$ & mean & p99 & mean & p99 & mean & p99 \\
\midrule
HiSparse & 4 & 0.79 & 1.50 & 0.70 & 0.95 & 19.4 & 19.5 \\
 & 8 & 1.05 & 2.81 & 0.75 & 1.82 & 23.4 & 23.6 \\
 & 16 & 1.85 & 5.49 & 1.20 & 3.49 & 28.4 & 28.7 \\
 & 32 & 3.82 & 10.81 & 2.49 & 7.40 & 34.8 & 36.4 \\
 & 64 & 5.09 & 21.35 & 2.68 & 9.63 & 42.2 & 43.4 \\
 & 128 & 10.68 & 43.53 & 5.11 & 26.97 & 50.3 & 51.8 \\
 & 256 & 47.61 & 158.86 & 42.54 & 102.89 & 58.0 & 59.1 \\
\addlinespace
Adaptive HiSparse & 4 & 0.54 & 0.79 & 0.48 & 0.71 & 16.0 & 16.1 \\
 & 8 & 0.76 & 1.07 & 0.69 & 0.97 & 19.1 & 19.2 \\
 & 16 & 1.24 & 3.14 & 0.90 & 1.23 & 24.1 & 24.7 \\
 & 32 & 1.59 & 5.70 & 1.25 & 3.09 & 30.8 & 31.7 \\
 & 64 & 3.47 & 16.74 & 2.05 & 6.66 & 38.1 & 39.9 \\
 & 128 & 6.13 & 36.16 & 2.48 & 11.85 & 46.6 & 48.3 \\
 & 256 & 34.36 & 131.61 & 31.01 & 81.99 & 54.4 & 55.4 \\
\addlinespace
Adaptive DHiSparse & 4 & 0.53 & 0.79 & 0.47 & 0.70 & 11.4 & 13.2 \\
 & 8 & 0.65 & 1.35 & 0.49 & 0.72 & 13.3 & 16.4 \\
 & 16 & 0.86 & 2.55 & 0.64 & 1.03 & 16.2 & 18.8 \\
 & 32 & 1.47 & 6.88 & 0.74 & 1.40 & 19.4 & 24.4 \\
 & 64 & 2.76 & 15.45 & 1.12 & 3.02 & 26.2 & 31.8 \\
 & 128 & 5.13 & 32.42 & 2.25 & 9.67 & 35.9 & 42.1 \\
 & 256 & 64.62 & 111.89 & 67.79 & 87.60 & 38.9 & 48.3 \\
\bottomrule
\end{tabular}
\end{minipage}\par
\endgroup
\vspace{2pt}

\endgroup
\end{document}